\documentclass{article}

\usepackage{microtype}
\usepackage{graphicx}
\usepackage{subcaption}
\usepackage{booktabs} % for professional tables

\usepackage{arydshln} % for dashed lines in tables

\usepackage{dblfloatfix}

\usepackage{tabularx,makecell,booktabs,pifont}
\newcommand{\cmark}{\ding{51}} % check
\newcommand{\xmark}{\ding{55}} % cross

\usepackage{hyperref}
\usepackage{arydshln}

\usepackage[most]{tcolorbox}

\newtcbox{\myinlinecode}{
    on line, 
    arc=3pt,                         % Adjust for more/less rounding
    boxrule=0pt,                      % No actual border line
    colback=gray!15,                  % Light gray background (15% black)
    colframe=gray!15,                 % Match border to background
    before upper=\rule[-0.2ex]{0pt}{0pt}\texttt, % Force monospace font
    boxsep=0pt, 
    left=3pt, 
    right=3pt, 
    top=2pt, 
    bottom=2pt
}

\usepackage[preprint]{icml2026}

\usepackage{amsmath}
\usepackage{amssymb}
\usepackage{mathtools}
\usepackage{amsthm}
\usepackage{tcolorbox}

\usepackage{makecell}
\usepackage{tabularx}
\usepackage{amssymb}
\usepackage{graphicx}

\usepackage{fontawesome5}
\usepackage{url}

\usepackage[capitalize,noabbrev]{cleveref}

\theoremstyle{plain}

\theoremstyle{definition}

\theoremstyle{remark}

\usepackage[textsize=tiny]{todonotes}

\icmltitlerunning{MATRIX: A Multimodal Benchmark and Post-Training Framework for Materials Science}

\usepackage{eso-pic}
\newcommand{\stampfirstpagelogo}{%
  \AddToShipoutPictureFG*{%
    \AtPageUpperLeft{%
      \raisebox{-10mm}{\hspace{8mm}\includegraphics[height=0.3cm]{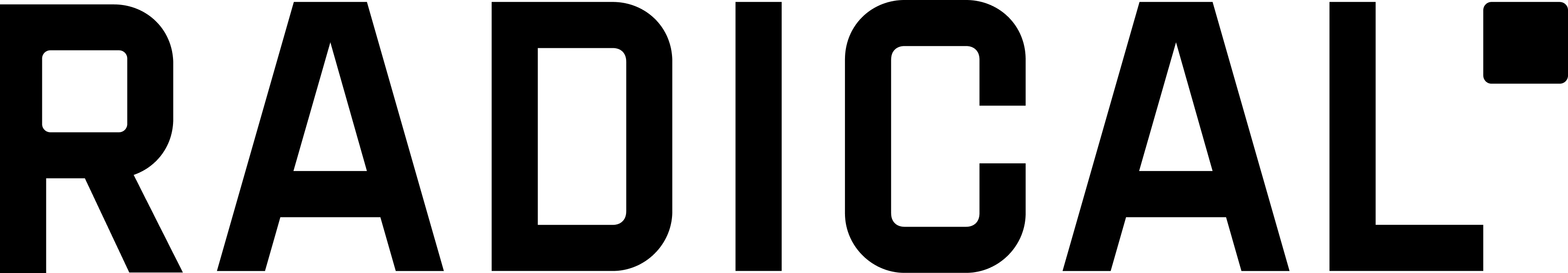}}%
    }%
  }%
}

\begin{document}
\stampfirstpagelogo
    \twocolumn[
    \icmltitle{MATRIX: A Multimodal Benchmark and Post-Training Framework for Materials Science}



  \icmlsetsymbol{equal}{*}

  \begin{icmlauthorlist}
    % \icmlauthor{Anonymous Author}{anon}
    \icmlauthor{Delia McGrath}{radical}
    \icmlauthor{Curtis Chong}{radical}
    \icmlauthor{Rohil Kulkarni}{radical}
    \icmlauthor{Gerbrand Ceder}{radical}
    \icmlauthor{Adeesh Kolluru}{radical,n}

  \end{icmlauthorlist}

  \icmlaffiliation{radical}{Radical AI}
  \icmlaffiliation{n}{Work done while at Radical AI}
  \icmlcorrespondingauthor{Delia McGrath}{dmcgrath@radical-ai.com}
    \icmlcorrespondingauthor{Adeesh Kolluru}{kolluru.adeesh@gmail.com}

\icmlkeywords{Machine Learning, AI for Science, Foundation Models, Multimodal Learning, Benchmarking}

  \vskip 0.3in
]



\printAffiliationsAndNotice{}  % no special notice (required even if empty)

\begin{abstract} % 215 words
Scientific reasoning in materials science requires integrating multimodal experimental evidence with underlying physical theory. Existing benchmarks make it difficult to assess whether incorporating visual experimental data during post-training improves mechanism-grounded explanation reasoning beyond text-only supervision. We introduce MATRIX, a multimodal benchmark for materials science reasoning that evaluates foundational theory, research-level reasoning, and the interpretation of real experimental artifacts across multiple characterization modalities. Using MATRIX as a controlled diagnostic, we isolate the effect of visual grounding by comparing post-training on structured materials science text alone with post-training that incorporates paired experimental images. Despite using relatively small amounts of multimodal data, visual supervision improves experimental interpretation by 10–25\% and yields 5–16\% gains on text-only scientific reasoning tasks. Our results demonstrate that these improvements rely on correct image–text alignment during post-training, highlighting cross-modal representational transfer. We also observe consistent improvements on ScienceQA and PubMedQA, demonstrating that the benefits of structured multimodal post-training extend beyond materials science. The MATRIX dataset is available at \href{https://huggingface.co/datasets/radical-ai/MATRIX}{hf.co/datasets/MATRIX} and model at {\href{https://huggingface.co/radical-ai/MATRIX-PT}{hf.co/models/MATRIX-PT}}.

% \centering
% \small
% \faIcon{database} \texttt{\href{https://huggingface.co/datasets/radical-ai/MATRIX}{hf.co/datasets/MATRIX}} \\
% \quad
% \faIcon{robot} \texttt{\href{https://huggingface.co/radical-ai/MATRIX-PT}{hf.co/models/MATRIX-PT}}

\end{abstract}

% The primary reason lies in the fact that most current MLLMs rely heavily on statistical pattern prediction, lacking explicit reasoning mechanisms.

\begin{table*}[t]
\centering
\footnotesize
\setlength{\tabcolsep}{3pt}
\renewcommand{\arraystretch}{1.15}

\begin{tabularx}{\textwidth}{l*{7}{>{\centering\arraybackslash}X}}
\toprule
Benchmark &
\makecell{Research\\reasoning} &
\makecell{Theory} &
\makecell{Hypothesis\\generation} &
\makecell{XRD} &
\makecell{EDS} &
\makecell{SEM} &
\makecell{TGA} \\
\midrule
ChemBench*~\cite{mirza2024large}
& \xmark & \cmark & \xmark & \xmark & \xmark & \xmark & \xmark \\
FrontierScience*~\cite{wang2025frontierscience}
& \cmark & \cmark & \xmark & \cmark & \xmark & \xmark & \xmark \\
MacBench~\cite{alampara2024macbench}
& \xmark & \xmark & \xmark & \cmark & \xmark & \xmark & \xmark \\
MatBookQA~\cite{mishra2024foundational}
& \xmark & \cmark & \xmark & \xmark & \xmark & \xmark & \xmark \\
MatSciBench~\cite{Zhang2025MatSciBench}
& \cmark & \cmark & \xmark & \xmark & \xmark & \xmark & \xmark \\
MaScQA~\cite{zaki2024mascqa}
& \xmark & \cmark & \xmark & \xmark & \xmark & \xmark & \xmark \\
\textbf{MATRIX (ours)}
& \cmark & \cmark & \cmark & \cmark & \cmark & \cmark & \cmark \\
\bottomrule
\end{tabularx}

\caption{Comparison of materials science benchmarks across reasoning level, hypothesis generation, and experimental modalities.
\protect\\ \footnotesize * Mixed-domain benchmark; only material-science questions were considered when determining coverage.}
\label{tab:benchmark_comparison}
\end{table*}

\section{Introduction}

Large language models (LLMs) have shown rapid progress in multi-step reasoning and structured problem solving~\cite{wei2022chain,wang2024chain}. As a result, foundation models are increasingly applied to scientific tasks such as analysis, hypothesis generation, and literature synthesis~\cite{wang2025drsr,prabhakar2025omniscience}. To track this progress, several benchmarks have been proposed to evaluate scientific reasoning and discovery capabilities~\cite{phan2025humanity,wang2025frontierscience}. However, most of these benchmarks mainly emphasize textual understanding of scientific concepts and explanations.

Particularly within materials science, scientific reasoning goes beyond text alone. Conclusions are grounded in experimental characterization, including SEM micrographs, XRD patterns, EDS spectra, and TGA curves~\cite{liu2023autonomous,sanou2025metrology}. Interpreting these measurements requires mapping perceptual structure—such as morphology, peak structure, contrast, and trends—onto physical concepts, including phases, microstructure, composition, and kinetics. Despite this reliance on experimental evidence, existing materials science benchmarks provide limited support for evaluating such reasoning: many focus on text-only questions~\cite{Zhang2025MatSciBench} or rely on multiple-choice and short-answer formats~\cite{alampara2024macbench}, which do not test whether models can integrate experimental evidence with theory or isolate whether gains on multimodal tasks arise from genuine visual grounding rather than stronger language modeling induced by post-training.

This distinction has become increasingly important as multimodal foundation models continue to improve. Recent analyses show that vision–language models can exhibit systematic compositional and binding failures even when surface-level perception appears strong~\cite{ma2023crepe,aravindan2025vlms}. At the same time, prior work in adjacent scientific domains shows that visual representations can support discovery and symbolic inference~\cite{liu2025mimicking,liu2025can}. Together, these findings raise a key open question: when multimodal foundation models are trained with experimental imagery, do observed gains reflect genuine improvements in scientific reasoning, or merely stronger perception when images are present?

% \subsection{Key Contributions}
% %Together, these findings raise two closely related questions.
% Within our work, we first aim to find how well current foundation models perform across textual and image-based materials science tasks. We then go on to find whether , do observed gains reflect deeper changes in scientific reasoning behavior or merely improved perception when images are present?
 Within our work, we first aim to find how well current foundation models perform across textual and image-based materials science tasks. We then explore the factors that improve reasoning for materials and how this generalizes across other scientific domains. 

We make the following contributions:
\begin{enumerate}
    \item Introduce MATRIX, a multimodal benchmark for scientific reasoning in materials science and its corresponding MATRIX-PT post-training dataset.
    \item We isolate the effects of post-training and demonstrate that aligned image–description pairs yield consistent gains in text-only reasoning on out-of-distribution tasks.
    \item Demonstrate that post-training on foundational and applied vision and textual materials science tasks has transferable gains on other science benchmarks.
\end{enumerate}

\section{MATRIX Benchmark}

\begin{figure}[t]
    \centering
    \includegraphics[width=.9\columnwidth]{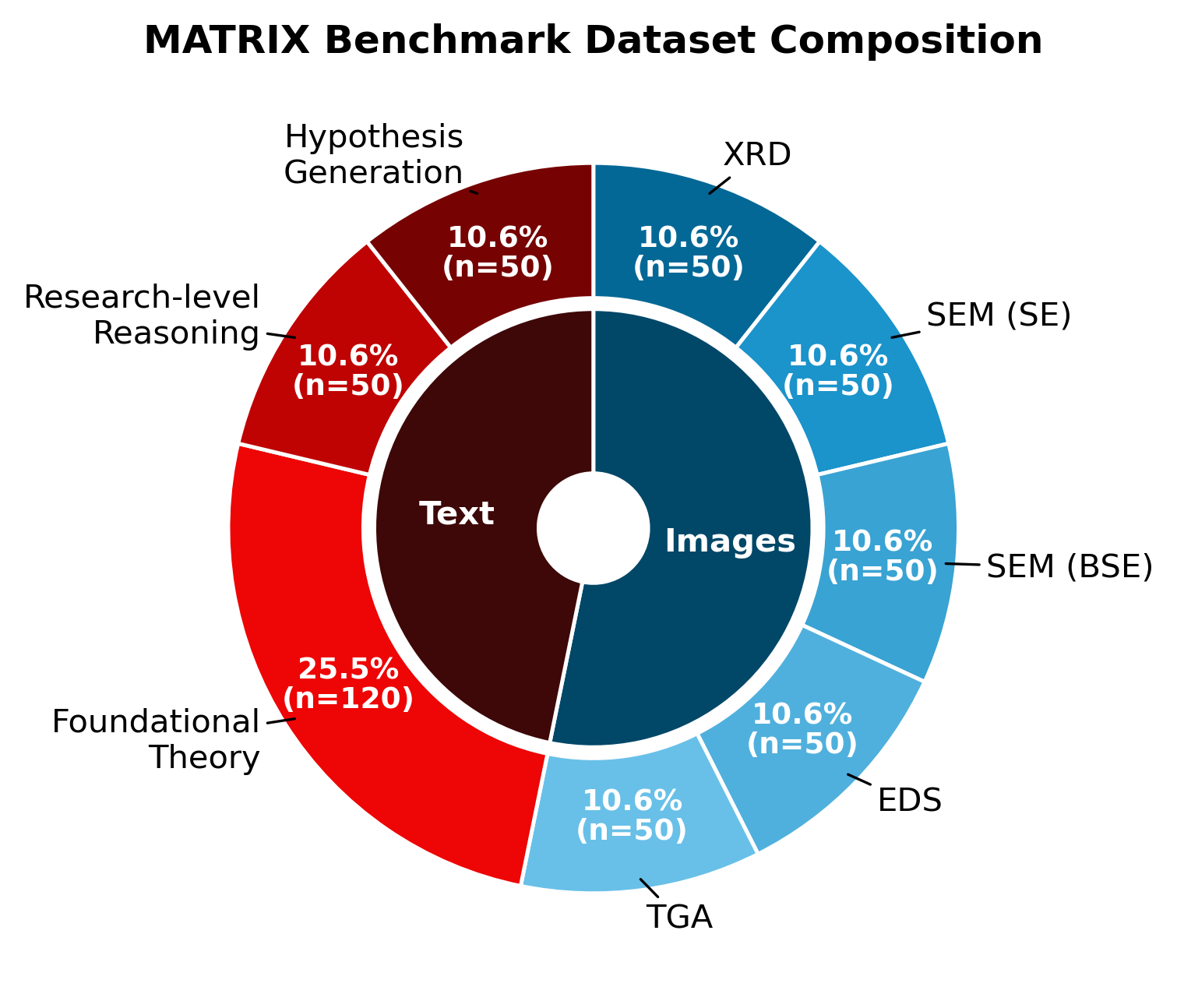}
    \caption{Breakdown of MATRIX benchmark by text and image tasks, further broken down by subtasks.}
    \label{fig:MATRIX_subsets}
\end{figure}

\begin{figure*}[t]
    \centering
    \includegraphics[width=.9\textwidth]{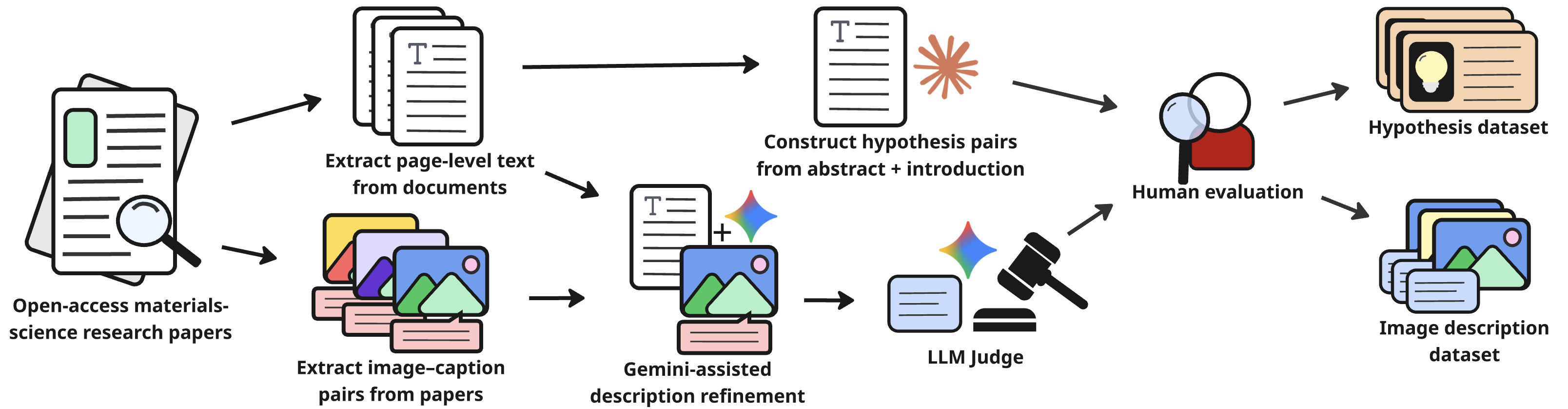}
    \caption{
MATRIX data curation and task construction pipeline.
Research papers are parsed into page-level text and experimental figure streams.
Limited early-paper context is used to construct hypothesis-generation prompts, while experimental figures (SEM-BSE, SEM-SE, XRD, EDS, TGA) are combined with their local textual context to construct image–caption pairs.
These pairs are standardized to support standalone experimental reasoning without reliance on full-paper memorization.
Validation and quality-control steps ensure that captions remain grounded in observable evidence.
See~\autoref{sec:appendix_data} for full implementation details and filtering criteria.
}
    \label{fig:MATRIX_research_pipeline}
\end{figure*}

% \subsection{Overview}

MATRIX (\textbf{M}aterials \textbf{A}nalysis for \textbf{T}heory, \textbf{R}easoning, and \textbf{I}mages from e\textbf{X}periments) is a multimodal benchmark designed to evaluate scientific reasoning in materials science. Rather than treating text-based reasoning and experimental interpretation as separate capabilities, it evaluates both within a single framework, enabling direct comparison across task families and modalities.

The benchmark comprises four task families—foundational theory reasoning, research-level reasoning, hypothesis generation, and experimental interpretation—constructed from open-access research papers and postgraduate-level coursework. Figure~\ref{fig:MATRIX_subsets} illustrates the distribution of text and image-based tasks across these families. Table~\ref{tab:benchmark_comparison} situates MATRIX relative to existing materials science benchmarks, highlighting differences in reasoning level, modality coverage, and task scope.

\subsection{Task Design}

MATRIX tasks are constructed from two complementary data sources: postgraduate-level materials science coursework and open-access research papers. Coursework provides structured text-based reasoning problems, including foundational theory and research-level reasoning tasks. While research papers provide experimental figures and partial textual context used to construct experimental interpretation and hypothesis generation tasks.

% MATRIX evaluates scientific reasoning across four task families that reflect how materials science understanding is developed in practice: (i) foundational theory reasoning, (ii) research-level reasoning, (iii) hypothesis generation, and (iv) experimental interpretation. Rather than isolating these competencies into separate benchmarks, MATRIX integrates them within a single framework to enable analysis of how textual and multimodal supervision jointly affect scientific reasoning behavior.

\subsubsection{Text-Based Reasoning Tasks}

Text-based reasoning tasks assess whether models can produce coherent, principle-driven explanations of fundamental and applied materials science concepts using language alone. These tasks are derived from postgraduate-level materials science coursework and include foundational theory questions and research-level reasoning problems. Unlike short-answer or multiple-choice formats, tasks require long-form explanatory responses that justify conclusions using underlying physical principles rather than recalling isolated facts~\cite{rueda2025understanding,zheng2023large}.

Foundational theory tasks assess conceptual understanding of core topics such as thermodynamics, phase equilibria, and structure--property relationships. Research-level reasoning tasks extend this foundation to open-ended scenarios adapted from graduate examinations, requiring multi-step qualitative inference, synthesis of multiple concepts, and explicit discussion of assumptions or trade-offs. Together, these tasks emphasize explanation-centered reasoning that mirrors how materials scientists articulate understanding in research contexts and discourages verbatim recall.

Because these tasks exclude visual inputs, performance on this subset provides a direct measure of language-based scientific reasoning and is used to assess cross-modal transfer effects from multimodal post-training. Representative examples and construction details are provided in Appendix~\ref{sec:appendix_data}.

\subsubsection{Hypothesis Generation}

Beyond explaining established concepts, scientific reasoning often requires proposing plausible next steps from an incomplete context. Hypothesis generation tasks assess whether models can propose hypotheses or follow-up investigations given early-stage information rather than fully specified outcomes.

Prior hypothesis-generation benchmarks typically measure the rigid recovery of ground-truth hypotheses in synthetic tasks or by framing generation as conditional modeling over structured problem--hypothesis pairs~\cite{liu2025hypobench,o2025sparks}. In contrast, MATRIX treats hypothesis generation as open-ended: we permit multiple acceptable responses, scoring them for scientific plausibility and consistency with the provided context rather than matching a single reference answer.

These tasks are constructed from early sections of open-access research papers (e.g., abstracts and introductions) to encourage forward-looking reasoning while avoiding leakage of reported results. Representative examples and evaluation criteria are provided in Appendix~\ref{sec:appendix_scoring}.

\begin{table*}[t]
\centering
\small
\setlength{\tabcolsep}{5pt}
\renewcommand{\arraystretch}{1.15}

\begin{tabular}{l ccccc | ccc}
\toprule
 & \multicolumn{5}{c|}{\textbf{Image-based Tasks}} & \multicolumn{3}{c}{\textbf{Text-based Tasks}} \\
\cmidrule(lr){2-6} \cmidrule(lr){7-9}
\textbf{Model}
& \textbf{EDS}
& \textbf{SEM-BSE}
& \textbf{SEM-SE}
& \textbf{TGA}
& \textbf{XRD}
& \textbf{Theory}
& \textbf{Hypothesis}
& \textbf{Research} \\
\midrule
\multicolumn{9}{l}{\textit{Open Source Vision-Language Models}} \\
InternVL2-8B 
& 0.345 & 0.350 & 0.375 & 0.430 & 0.430
& 0.402 & 0.510 & 0.505 \\
DeepSeek-VL-7B-chat
& 0.360 & 0.420 & 0.425 & 0.475 & 0.395
& 0.392 & 0.535 & 0.568 \\
Qwen2-VL-7B-Instruct
& 0.210  & 0.145 & 0.195 & 0.270 & 0.175 
              
  % - other: 0.215                                                                                               
& 0.385 & 0.495 & 0.555 \\
Qwen3-VL-8B-Instruct
& 0.385 & 0.400 & 0.415 & 0.445 & 0.320
& \textbf{0.600} & \textbf{0.765} & 0.695 \\

Ministral-3-8B-Instruct
& 0.475 & 0.480 & 0.485 & 0.545 & 0.49
& 0.589 & 0.685 & 0.685 \\

LLaMaT-2 & * &* &* &* &* & 0.427 & 0.540 & 0.530 \\
LLaMaT-3 & * &* &* &* &* & 0.400 & 0.550 & 0.525 \\

\midrule
\multicolumn{9}{l}{\textit{Base Model (Starting Point for Our Post-Training)}} \\
Qwen2-VL-7B-Base
& 0.455	& 0.395 & 	0.435	& 0.540	& 0.465
& 0.414 & 0.530 & 0.510 \\
\midrule
\multicolumn{9}{l}{\textit{Qwen Post-Trained Models (MATRIX-PT)}} \\
MATRIX-PT (Text)   & 0.520 & 0.455 & 0.470 & 0.520 & 0.485 & 0.470 & 0.580 & 0.565 \\
MATRIX-PT (Vision) & 0.580 & 0.580 & \textbf{0.610} & 0.590 & 0.610 & 0.450 & 0.605 & 0.635 \\
\textbf{MATRIX-PT (Full)} & \textbf{0.590} & \textbf{0.610} & 0.580 & \textbf{0.600} & \textbf{0.615} & 0.527 & 0.575 & \textbf{0.715} \\

\bottomrule
\end{tabular}

\caption{Performance on MATRIX across general-purpose vision-language models and our domain-adapted variants. \textbf{Image-based tasks} evaluate characterization data interpretation (EDS, SEM, TGA, XRD). \textbf{Text-based tasks} evaluate scientific reasoning without images. General-purpose models have not been adapted to materials science. Our domain-adapted models are trained on materials science papers and coursework. The Qwen2-VL-7B-Base model serves as the starting point for our post-training ablations. $^{*}$Text-only models}
\label{tab:model_comparison_full}
\end{table*}

\subsubsection{Experimental Interpretation}

Experimental interpretation tasks assess whether models can infer materials-relevant conclusions from real characterization data by combining visual evidence with domain knowledge. Unlike generic image captioning or object recognition, these tasks require mapping observed visual or graphical structure to underlying physical concepts such as phase composition, microstructure evolution, elemental distribution, and thermal behavior.

MATRIX includes experimental artifacts commonly used in materials research, including SEM micrographs acquired in both backscattered electron (BSE) and secondary electron (SE) modes, as well as XRD patterns, EDS spectra, and TGA curves. Tasks are constructed from open-access research papers using a structured curation and alignment pipeline (Figure~\ref{fig:MATRIX_research_pipeline}). Prompts include limited local textual context to supply essential experimental background—such as composition or measurement conditions—required for scientifically valid interpretation when such information is not directly observable from the signal.

Prior work has explored reasoning over scientific microscopy in other domains, particularly biology~\cite{lozano2024micro,burgess2025microvqa}. However, materials science interpretation poses distinct challenges: many characterization signals encode abstract physical quantities rather than discrete objects, and conclusions depend on understanding both the measurement modality and its experimental context. MATRIX reflects these challenges by requiring models to interpret modality-specific signals rather than surface-level visual patterns.

Additional construction details and representative examples are provided in Appendix~\ref{sec:appendix_data}. Results on experimental interpretation tasks are reported separately in Table~\ref{tab:model_comparison_full}, enabling direct comparison with text-only reasoning performance.

\subsection{Evaluation}

All MATRIX tasks require long-form explanatory responses. Consequently, submissions are evaluated using a standardized rubric based on scientific correctness, reasoning quality, and consistency with the provided context. We use \textit{GPT-5.1} as a rubric-guided LLM judge that assigns scores on a five-point scale $\{0, 0.25, 0.5, 0.75, 1.0\}$. Furthermore, rubrics are task-specific and account for the open-ended nature of hypothesis generation and experimental interpretation. Judge calibration and validation procedures are described in Appendix~\ref{sec:appendix_scoring}. Quantitative results for all evaluated models are summarized in Table~\ref{tab:model_comparison_full}.

\subsection{Results Against Open-Source Vision-Language Models}
We evaluate representative open-source vision–language models (instruction-tuned VLMs) alongside materials-specialized text-only post-trained baselines~\cite{mishra2024foundational}, all at roughly 7B parameters (top of~\autoref{tab:model_comparison_full}). We first describe baseline results, then analyze how domain-adapted supervision changes performance across task families.

\autoref{tab:model_comparison_full} separates performance on experimental interpretation (EDS, SEM, TGA, XRD) from text-based reasoning (theory, hypothesis generation, research reasoning). Across all models, text-based reasoning is substantially stronger than experimental interpretation. While post-trained models perform well on hypothesis generation and research reasoning, they consistently underperform on microscopy and diffraction-based tasks, indicating that strong language-based scientific reasoning does not automatically extend to vision-grounded materials interpretation. Notably, vision-only post-training improves performance on text-based reasoning tasks, indicating that multimodal supervision can reshape internal representations even when visual inputs are absent at inference time.

While joint text+vision post-training yields the best balance across tasks, it does not uniformly outperform modality-specific supervision on every subtask. In particular, text-only post-training remains competitive on narrowly scoped text-based reasoning tasks, and vision-only post-training excels on certain image-based modalities. This pattern reflects a tradeoff: multimodal grounding improves integrative reasoning across evidence types, but does not necessarily dominate specialized supervision on isolated subtasks.

Not all experimental modalities pose equal difficulty. Lower-dimensional signals such as TGA curves are handled comparatively well, while spatially and structurally complex modalities like SEM and XRD remain challenging. Across baselines in Table~\ref{tab:model_comparison_full}, TGA scores are generally higher than SEM and XRD, while microscopy and diffraction-based interpretation remain the most inconsistent. This suggests that current open-source VLMs capture some coarse structure in experimental signals, but struggle to reliably map modality-specific features to materials-relevant conclusions.

Text-only materials-specialized models remain competitive on theory and research reasoning despite lacking visual inputs. This aligns with prior findings that high-quality textbook-style instructional data strongly supports conceptual and explanatory reasoning in language models~\cite{gunasekar2023textbooks}. Since MATRIX emphasizes long-form mechanistic explanation over numerical derivation, such exposure confers a clear advantage on text-based subsets.

Overall, these results highlight a mismatch in existing open-source VLMs: strong text-only scientific reasoning alongside weak experimental interpretation. This gap motivates our controlled post-training experiments, which test whether aligned multimodal supervision can improve experimental reasoning and reshape text-only scientific performance.

As with other foundation models, some overlap with publicly available scientific literature cannot be ruled out; however, MATRIX evaluates structured reasoning behavior rather than memorization of specific results.

\section{Post-Training Configurations and Cross-Modal Transfer}

\label{sec:post_training}

We design post-training experiments to isolate how textual reasoning supervision and aligned visual grounding affect scientific reasoning. Our goal is diagnostic rather than performance-maximizing, following prior work on controlled evaluation of foundation models~\cite{bommasani2021opportunities}.

Post-training supervision spans foundational theory, research-level reasoning, hypothesis generation, and aligned experimental image--text pairs. We intentionally allocate a larger proportion of supervision to foundational theory, as textbook-style instructional data has been shown to provide a strong signal for conceptual and explanatory reasoning~\cite{gunasekar2023textbooks}. This ensures strong baseline language competence and reduces confounding effects when analyzing cross-modal transfer.

To avoid data contamination, post-training and evaluation data are disjoint at the document level. We refer to the post-training dataset as MATRIX-PT, which is constructed from documents disjoint from those used in the MATRIX benchmark and shares no overlapping figures, captions, or text. Further implementation details and data statistics are provided in~\autoref{sec:appendix_post_training}.

\subsection{Post-Training Conditions}
We evaluate four post-training conditions:
\begin{enumerate}
    \item \textbf{Base}: The unadapted Qwen2-VL-7B model.
    \item \textbf{Text-only SFT}: Supervised fine-tuning on structured materials science reasoning tasks (foundational theory, research reasoning, hypothesis generation).
    \item \textbf{Vision-only SFT}: Supervised fine-tuning on experimental image--caption pairs (SEM, XRD, EDS, TGA).
    \item \textbf{Text + Vision SFT}: Joint fine-tuning on the union of text-based reasoning tasks and experimental image--caption pairs.
\end{enumerate}

Structured text supervision has been shown to inject domain knowledge and improve reasoning in language models~\cite{gururangan2020don,mecklenburg2024injecting}, while multimodal representation learning enables shared abstractions across vision and language~\cite{radford2021learning,alayrac2022flamingo}. Our configurations are designed to disentangle these effects.

\begin{table*}[t]
\centering
\begin{small} % Slightly smaller font often looks better for wide tables
\addtolength{\tabcolsep}{-2pt} % Adjust horizontal spacing to fit margins
\begin{tabular}{l ccccc|ccc}
\toprule
& \multicolumn{5}{c}{\textbf{Image-based Tasks}} & \multicolumn{3}{c}{\textbf{Text-based Tasks}} \\
\cmidrule(lr){2-6} \cmidrule(lr){7-9}
\textbf{Alignment Disruption} & EDS & SEM-BSE & SEM-SE & TGA & XRD & Theory & Hypo. & Reason. \\
\midrule
\textit{MATRIX-PT (Full)} \\
Aligned (default) & \textbf{0.590} & \textbf{0.610} & \textbf{0.580} & \textbf{0.600} & \textbf{0.615} & \textbf{0.527} & \textbf{0.575} & \textbf{0.715} \\
\addlinespace[0.5em]
Answer permutation & 0.010 & 0.000 & 0.000 & 0.010 & 0.005 & 0.506 & 0.515 & 0.560 \\
Image permutation  & 0.260 & 0.235 & 0.255 & 0.400 & 0.335 & 0.502 & 0.555 & 0.660 \\
Image removal      & 0.230 & 0.220 & 0.220 & 0.340 & 0.305 & 0.519 & 0.525 & 0.625 \\
\bottomrule
\end{tabular}
\end{small}
\caption{
Effect of disrupting image--text alignment during multimodal post-training.
Each row corresponds to a controlled permutation of the question--image--answer structure illustrated in Figure~\ref{fig:vision_ablation}.
Disrupting alignment severely degrades performance on image-based experimental interpretation tasks, while text-based reasoning remains comparatively stable.
}
\label{tab:caption-performance}

\end{table*}

\subsection{Cross-Modal Representational Transfer and Scientific Reasoning}

Recent work on vision--language models (VLMs) has demonstrated strong performance on perception-driven tasks, including captioning, visual question answering, and domain-specific image interpretation in settings such as microscopy and medical imaging~\cite{verma2024beyond,chen2025think}. However, these evaluations predominantly assess perceptual accuracy or localized reasoning. They do not test whether visual grounding alters the abstract scientific representations that support higher-level reasoning.

At the same time, a growing body of work has shown that VLMs exhibit systematic failures in compositional generalization and mechanistic reasoning, even when perceptual performance appears strong~\cite{pearson2025evaluating,aravindan2025vlms}. These findings suggest that perception alone is insufficient for robust reasoning and that the interaction between perceptual grounding and symbolic or mechanistic abstraction remains poorly understood. In scientific domains, this gap is particularly salient: interpreting experimental artifacts such as SEM micrographs or XRD patterns requires mapping visual structure onto latent physical concepts that are rarely fully specified by text alone.

We hypothesize that multimodal post-training improves scientific reasoning by reshaping shared latent representations that underlie both visual perception and language-based abstraction. In materials science, experimental interpretation requires aligning domain-specific language (e.g., ``peak broadening,'' ``lamellar morphology,'' or ``secondary phase segregation'') with latent physical structure inferred from images. This alignment pressure may encourage representations that better capture mechanistic regularities, which are then reused in text-only reasoning tasks such as hypothesis generation or research-level explanation.

This view is consistent with prior arguments that grounding perceptual signals can refine abstract representations rather than merely adding a separate perceptual skill~\cite{bisk2020experience,alayrac2022flamingo}. More recent work has further suggested that structured or iterative multimodal reasoning can lead to improved abstraction and generalization~\cite{sharma2025see,liu2025mimicking}, though these effects are rarely evaluated outside of perception-centric tasks.

Formally, let \( p(x^{(v)}, x^{(t)}, z) \) model the joint distribution, where the modalities
\( x^{(v)} \) (images) and \( x^{(t)} \) (text) are generated from a shared latent variable \( z \):
\[
\begin{aligned}
p(x^{(v)}, x^{(t)}, z)
&= p(z)\, p(x^{(v)} \mid x^{(t)}, z)\, p(x^{(t)} \mid z) \\
&= p(z)\, p(x^{(v)} \mid z)\, p(x^{(t)} \mid z).
\end{aligned}
\]
{\footnotesize Where $p(x^{(v)} \mid x^{(t)}, z) = p(x^{(v)} \mid z)$ by the conditional independence of $x^{(v)}$ and $x^{(t)}$ given $z$.}

Training encoders $f_v(x^{(v)}) \approx z$ and $f_t(x^{(t)}) \approx z$ on paired data enables cross-modal transfer: improvements in $f_v$ from image supervision refine the shared representation $z$, benefiting text-only reasoning downstream~\cite{radford2021learning}.

MATRIX is explicitly designed to make this effect observable. Because text-based tasks are evaluated without visual input at inference time, any improvement from multimodal post-training reflects representational transfer rather than test-time access to images. Existing benchmarks, which either evaluate perception in isolation or lack structured scientific reasoning tasks, cannot surface this phenomenon.

\subsection{Effect of Multimodal Post-Training on Text-Only Scientific Reasoning}

\autoref{tab:model_comparison_full} summarizes MATRIX performance across post-training configurations. Text-only supervised fine-tuning yields clear improvements over the base model on foundational theory and research-level reasoning, confirming that structured domain supervision strengthens scientific language modeling.

Particularly, adding multimodal experimental grounding yields further gains not only on image-based interpretation tasks, but also on \emph{text-only} reasoning categories at inference time. These improvements are most pronounced on tasks that require multi-step explanation and mechanistic reasoning, such as research-level and long-answer questions.

Performance gains on experimental modalities are largest for SEM tasks, where microstructural complexity and underspecified captions place greater demands on visual–theoretical integration. While text-only fine-tuning produces modest improvements in image caption quality, multimodal fine-tuning substantially improves reasoning over experimental evidence.

Together, these results provide direct evidence of cross-modal transfer: visual grounding alters internal representations that are reused by text-only scientific reasoning. This effect is not detectable using existing benchmarks that evaluate text and images in isolation.

\begin{figure}[t]
    \centering
    \includegraphics[width=.58\columnwidth]{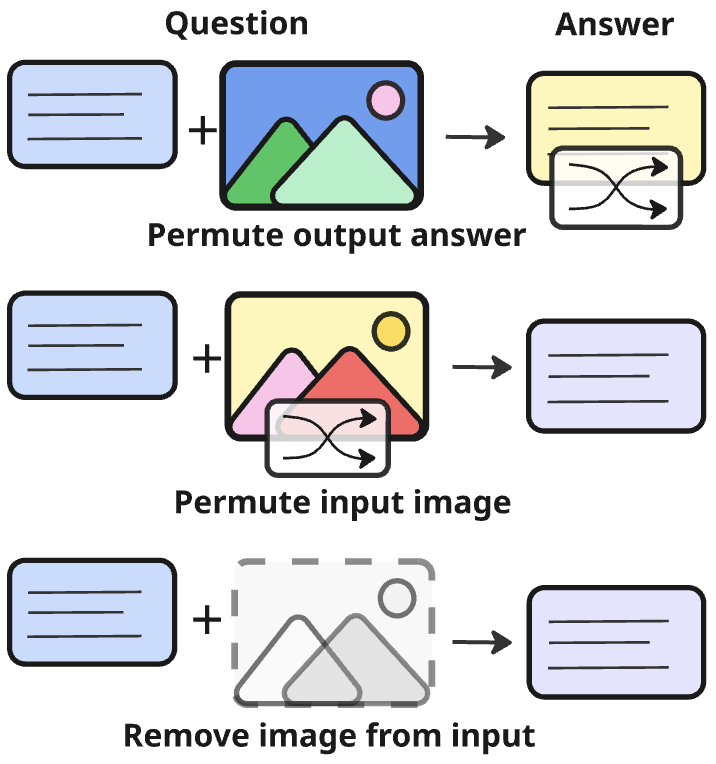}
    \caption{Each row illustrates a controlled modification of the question–image–answer structure used during post-training.
Top: The answer text is permuted while the question and image remain fixed.
Middle: The input image is permuted while the question and answer text remain fixed.
Bottom: The image is removed entirely.
These controls test whether observed gains arise from correct image–text alignment rather than spurious correlations or answer memorization.}
\label{fig:vision_ablation}
\end{figure}
\section{Further Exploration of Visual Grounding Effects}

\begin{table*}[t]
\centering
\small
\setlength{\tabcolsep}{6pt}
\renewcommand{\arraystretch}{1.15}
\begin{tabular}{l ccc|cc}
\toprule
& \multicolumn{3}{c|}{\textbf{Text-only Evaluation}}
& \multicolumn{2}{c}{\textbf{Text + Vision Evaluation}} \\
\cmidrule(lr){2-4} \cmidrule(lr){5-6}
\textbf{Model}
& \textbf{ScienceQA-T}
& \textbf{PubMedQA-T}
& \textbf{SciBench}
& \textbf{ScienceQA}
& \textbf{SciBench} \\
\midrule
LLaMaT-2
& 61.83 & 64.3 & 2.60
& \multicolumn{2}{c}{--} \\
LLaMaT-3
& 46.18 & 64.9 & 4.19
& \multicolumn{2}{c}{--} \\
Qwen2-VL-7B-Base
& 70.41 & 43.5 & 9.83
& 71.15 & 10.20 \\
\midrule
%MATRIX-PT (Vision only) & 74.63 &  67.50 & \textbf{-}
%& 79.27 & \textbf{-} \\'

\textbf{MATRIX-PT (Ours)} & \textbf{79.37} & \textbf{71.8} & \textbf{10.26}
& \textbf{79.49} & \textbf{10.44} \\
\bottomrule
\end{tabular}
\caption{
Performance on scientific benchmarks under text-only and text+vision evaluation.
Text-only results for ScienceQA~\cite{saikh2022scienceqa} and PubMedQA~\cite{jin2019pubmedqa} exclude vision-dependent questions, while text+vision results are reported using the full benchmarks.
SciBench~\cite{wang2023scibench} is text-only by design. ScienceQA results were evaluated without hints. LLaMaT‑2 and LLaMaT‑3 are text-only models; therefore, results for text+vision benchmarks are not applicable (indicated by ‘--’).
}
\label{tab:generalise_transfer_table}
\end{table*}

\subsection{Motivation and Setup}

Results on MATRIX indicate that multimodal post-training improves not only experimental interpretation but also text-only scientific reasoning. However, improvements in text-based reasoning could arise from alternative mechanisms unrelated to visual grounding, such as exposure to additional domain-specific text during image–caption supervision, implicit answer memorization, or spurious correlations between images and answers~\cite{bisk2020experience,alayrac2022flamingo}.

To disentangle these possibilities, we design a set of controlled perturbations to the question–image–answer structure used during multimodal post-training. These controls preserve the overall training format and supervision volume while selectively disrupting image–text alignment, allowing us to isolate the role of correct visual grounding. Each perturbation targets a distinct failure mode that could otherwise explain observed gains.

As illustrated in~\autoref{fig:vision_ablation}, we consider three variants. In \textbf{Answer Permutation}, the answer text is permuted across samples while the question and image remain fixed, breaking the correspondence between visual evidence and explanatory supervision. In \textbf{Image Permutation}, the input image is permuted while the question and answer text remain fixed, preserving the textual reasoning signal while removing meaningful visual grounding. In \textbf{Image Removal}, the image is removed entirely from the input, isolating the effect of text-only supervision within the multimodal training format. Details for these experiments are in~\autoref{sec:appendix_permutation}.

We further explored the effects of isolating the effect of pure text finetuning on each of the individual training sets for theory, research reasoning, and foundational theory, which are included in~\autoref{sec:appendix_add_res}~\autoref{tab:text_only_comparison}.

\subsection{Results}

\autoref{tab:caption-performance} reports performance across image-based experimental interpretation tasks and text-based reasoning tasks for each control condition. All three variants underperform the fully aligned Text+Vision post-training configuration, indicating that correct image–text alignment is necessary to realize the full benefits of multimodal supervision.

Disrupting the correspondence between images and answer text leads to severe degradation on image-based tasks. In the \textbf{Answer Permutation} condition, performance collapses across all experimental modalities, with near-zero scores on SEM, XRD, and EDS tasks. This result confirms that gains in experimental interpretation are not driven by superficial visual features or generic captions, but depend on aligned explanatory supervision.

Permuting the image while preserving the correct answer text (\textbf{Image Permutation}) substantially degrades experimental interpretation performance relative to the aligned model, with large drops observed across SEM, XRD, and TGA tasks. Notably, vision-only post-training improves performance on text-only tasks such as hypothesis generation (Table~\ref{tab:model_comparison_full}), providing direct evidence of cross-modal transfer despite images being absent at inference time.

Removing the image entirely (\textbf{Image Removal}) produces a similar pattern. Text-based reasoning performance remains modestly strong, while image-based task performance is consistently lower than in the aligned Text+Vision setting. The similarity between the \textbf{Image Permutation} and \textbf{Image Removal} conditions suggests that unaligned or irrelevant visual input does not contribute meaningfully to learning experimental abstractions.

Across all three controls, text-based reasoning performance remains relatively stable compared to image-based tasks, indicating that the observed gains in text-only scientific reasoning under proper multimodal post-training are not explained by spurious image–answer correlations or memorization of domain text. Instead, they appear to depend on aligned visual grounding during training.

\subsection{Interpretation}

Together, these results support a representation-level explanation for the gains observed under multimodal post-training. Correct alignment between experimental imagery and explanatory text appears essential for shaping internal representations that support both experimental interpretation and downstream text-only scientific reasoning. When this alignment is disrupted, models fail to learn abstractions that connect visual structure to latent physical concepts, even when comparable textual supervision is present.

These findings are consistent with prior arguments that perceptual grounding can refine abstract representations rather than merely adding a separate perceptual skill~\cite{bisk2020experience,radford2021learning}. In materials science, where experimental evidence plays a central role in mechanistic explanation, such grounded representations appear critical for robust scientific reasoning across modalities.

\section{Transfer Gains on Other Scientific Benchmarks}

To assess whether improvements induced by multimodal post-training extend beyond the MATRIX benchmark, we evaluate the post-trained model on a set of established scientific reasoning benchmarks spanning different domains and task formats: ScienceQA~\cite{saikh2022scienceqa}, PubMedQA~\cite{jin2019pubmedqa}, and SciBench~\cite{wang2023scibench}. We additionally compare against LLaMaT, a materials science–focused post-trained model, to distinguish gains arising from reasoning supervision from those attributable to domain-specific stylistic adaptation.

As shown in~\autoref{tab:generalise_transfer_table}, MATRIX post-training yields consistent improvements over both the base model and LLaMaT across benchmarks. Gains on ScienceQA are expected, as the benchmark emphasizes general scientific explanation and qualitative reasoning. More notably, the largest relative improvement is observed on PubMedQA, a biomedical benchmark constructed from PubMed abstracts. Despite the absence of biological data during post-training, the model exhibits substantially stronger performance, suggesting that exposure to graduate- and research-level scientific reasoning tasks enhances transferable abstract reasoning capabilities that generalize across scientific domains.

Performance on SciBench reveals a more nuanced pattern. While overall gains are modest, closer inspection indicates that improvements are concentrated in tasks that emphasize conceptual understanding and qualitative scientific reasoning. In contrast, tasks requiring explicit mathematical derivations—such as those drawn from postgraduate foundation problems with a single ground-truth solution—remain challenging. This distinction highlights a key boundary of transfer: multimodal post-training primarily benefits applied and open-ended scientific reasoning rather than symbolic or formula-driven problem solving.

Together, these results suggest that the benefits of MATRIX post-training extend beyond materials science but are selective in nature. Multimodal grounding and research-oriented supervision appear to strengthen abstract, explanation-driven scientific reasoning, while fully formal derivational competence likely requires more direct exposure to textbook-style mathematical instruction.

\section{Conclusion}

We introduce MATRIX, a multimodal benchmark for materials science that integrates reasoning over theory, textual knowledge, hypothesis generation, and real-world experimental images. By grounding evaluation in microscopy and laboratory data drawn from real experimental settings, MATRIX enables systematic study of vision-based scientific reasoning under realistic conditions.

We find that existing open-source multimodal models struggle on vision-grounded reasoning tasks in materials science. However, targeted post-training on MATRIX yields consistent improvements in both experimental interpretation and text-only scientific reasoning. Through controlled perturbations of image--text alignment, we also isolate the contribution of visual grounding and demonstrate that this alignment is responsible for these gains.

Models post-trained on our dataset also showed effective transfer across domains to established scientific reasoning benchmarks, including biomedical question answering. Analysis across benchmarks indicates that these gains are selective: multimodal grounding and research-oriented supervision primarily strengthen applied, explanation-driven scientific reasoning, while tasks requiring formal mathematical derivation remain challenging.

Taken together, our results suggest that multimodal grounding reshapes internal representations by aligning domain-specific language with latent physical structure, supporting mechanistic abstraction that can be reused across modalities. These findings reveal that benchmarks and training curricula grounded in real experimental evidence will play a central role in advancing generalizable scientific reasoning capabilities in future foundation models.

\section{Impact Statement}

This work introduces MATRIX, an open benchmark for evaluating multimodal scientific reasoning in materials science. By emphasizing long-form, mechanism-grounded explanations over short-form or perception-only tasks, MATRIX reflects how materials scientists integrate theoretical understanding with experimental evidence in real research workflows. We release the complete benchmark, including task formulations, evaluation protocols, and baseline implementations, to enable reproducible assessment of multimodal foundation models on scientific reasoning tasks.

The benchmark is intended to support the development and analysis of AI systems that assist with materials characterization, experimental interpretation, and research synthesis, where coherent mechanistic reasoning and careful use of empirical evidence are essential. MATRIX spans diverse experimental modalities, including SEM, XRD, EDS, and TGA, which are central to modern materials research. While the benchmark is grounded in materials science, its design principles are applicable to other scientific domains that require integrating visual evidence with abstract reasoning.

MATRIX evaluates reasoning quality using a rubric-based assessment of explanatory responses and does not replace experimental validation, domain expertise, or physical modeling. Performance on the benchmark reflects a model’s ability to articulate plausible scientific reasoning rather than empirical correctness or predictive accuracy in laboratory settings. As such, MATRIX should be used as a research and evaluation tool rather than as a basis for autonomous scientific decision-making. We anticipate that MATRIX will facilitate more systematic study of multimodal reasoning in scientific AI systems while highlighting the importance of grounded, mechanism-driven evaluation in technical domains.

\bibliography{refs}
\bibliographystyle{icml2026}

%%%%%%%%%%%%%%%%%%%%%%%%%%%%%%%%%%%%%%%%%%%%%%%%%%%%%%%%%%%%%%%%%%%%%%%%%%%%%%%
%%%%%%%%%%%%%%%%%%%%%%%%%%%%%%%%%%%%%%%%%%%%%%%%%%%%%%%%%%%%%%%%%%%%%%%%%%%%%%%
% APPENDIX
%%%%%%%%%%%%%%%%%%%%%%%%%%%%%%%%%%%%%%%%%%%%%%%%%%%%%%%%%%%%%%%%%%%%%%%%%%%%%%%
%%%%%%%%%%%%%%%%%%%%%%%%%%%%%%%%%%%%%%%%%%%%%%%%%%%%%%%%%%%%%%%%%%%%%%%%%%%%%%%

\newpage
\appendix

% \section{Appendix Overview}
% This appendix is intentionally where ``word-vomit'' belongs for ICML: full pipeline details, parsing ablations, filtering heuristics, prompt schemas, and evaluation rubrics.

\section{Data Processing Details}
\label{sec:appendix_data}

\subsection{Source Data}
\label{sec:appendix_source_data}
MATRIX draws text-based tasks from \emph{postgraduate core materials science courses}, including lecture notes/slides and examination questions. These materials provide structured coverage of foundational concepts and are written to elicit long-form explanation and synthesis. We aggregate multiple years of the same courses to preserve curricular coherence while introducing variation in formulation and difficulty.

MATRIX also draws from open-access materials science research papers. These papers provide contexts for hypothesis-generation tasks and supply experimental figures for image-based reasoning. For each paper, we extract figures and captions and retain surrounding page-level text where the figure is discussed, since characterization evidence is typically interpreted jointly with local scientific context.

\subsection{PDF Text Extraction}
\label{sec:appendix_text_extraction}

We extract text from PDFs using \texttt{pypdfium2}~\cite{pypdfium2}. We evaluated several widely used PDF parsing tools—PyPDF~\cite{pypdf}, Textract~\cite{textract_py}, Apache Tika~\cite{apache_tika}, and PyMuPDF~\cite{pymupdf}—on a pilot set of two-column scientific papers containing equations, chemical compositions, and Greek symbols. We selected \texttt{pypdfium2} due to its superior symbol fidelity, preservation of equation structure, and continuity of extracted text, consistent with prior comparative studies~\cite{adhikari2024comparative}.

After extraction, we apply a series of rule-based normalization steps to remove non-content artifacts, including page numbers, headers and footers, author affiliations, email addresses, copyright and license statements, and publisher boilerplate. Inline citations are removed using regular expressions to eliminate bracketed and parenthetical references, URLs, and DOI patterns, as citation formatting varies substantially across sources and introduces noise into downstream reasoning tasks.

We further normalize Unicode escapes and correct malformed whitespace and line breaks introduced during parsing. To standardize document scope, each paper is truncated to begin at the abstract and to end prior to the references section. While this procedure excludes appendices and supplementary material, we found that this restriction improves text consistency and reduces the inclusion of irrelevant names and citation lists, resulting in higher-fidelity inputs for reasoning-oriented tasks.

\subsection{Figure Extraction and Filtering}
\label{sec:appendix_figure_extraction}

Figures and captions from 63,000 open-access papers are extracted using Docling. Docling extracts all images and associated captions from each paper; figures without captions are discarded.

To select high-quality and descriptive image-caption pairs, we applied heuristic filters. First, captions must be at least 200 characters in length. Next, 80\% of the time, we filter out image-caption pairs if the figure contains multiple images. These are often low-quality captions because they describe multiple subfigures, rather than describing a single image in detail. We detect multi-figure captions by searching for subfigure identifiers (\myinlinecode{A)}, \myinlinecode{i)}). We kept subfigure images 20\% of the time to ensure our training distribution contained multi-image figures. We did not explore more sophisticated semantic methods to estimate caption quality, as we found that these filters were strict enough to constitute our dataset.

Once our initial set of captions was selected, we asked an LLM to look at each image-caption pair and categorize them into one of 7 types: \myinlinecode{TGA}, \myinlinecode{XRD}, \myinlinecode{EDS}, \myinlinecode{SEM-BSE}, \myinlinecode{SEM-SE}, \myinlinecode{other}, and \myinlinecode{irrelevant}. \myinlinecode{irrelevant} captions are those that do not describe the image at all (due to a Docling extraction error). Lastly, \myinlinecode{other} image-caption pairs are those that do not fit in the other five main categories.

To improve consistency across heterogeneous sources, we remove paper-specific information such as figure numbering (e.g., “Figure 3”). Instead, captions are treated as free-form descriptions, and papers are retained based on the presence of relevant domain keywords rather than caption structure. This design choice reflects the diversity of writing styles across journals and aims to standardize downstream reasoning inputs.

A central challenge in figure-based reasoning is the limited contextual information available when images are considered independently from the full paper. To address this, each image–caption pair is supplemented with a standardized context description derived from (i) the extracted image, (ii) the surrounding page-level text, and (iii) the original caption. This information is provided to a language model tasked with producing a concise, context-aware caption suitable for standalone interpretation.

To improve annotation reliability, a judge model independently evaluates whether the caption is plausibly inferable from the image and the provided context alone. This judge also ensures that the context does not spoil the answer by describing the image. Lastly, human inspection is applied to a subset of examples for quality control. This lengthy validation process reduces hallucinated details and ensures that generated captions reflect observable structure rather than unstated experimental assumptions.

Such rigorous standardization is essential for enabling benchmark usage outside the original publication context, including application to real laboratory images, where users may supply partial contextual information. By decoupling figure interpretation from paper-specific info, MATRIX supports evaluation of scientific reasoning grounded in experimental visual evidence rather than document memorization.

\section{Text-Based Theory and Research-Level Reasoning Tasks}
Foundational theory tasks assess conceptual understanding of core materials science topics, including crystallography, thermodynamics, phase transformations, and structure–property relationships. Questions are derived from postgraduate coursework and emphasize explanation and justification rather than factual recall.

Research-level reasoning tasks extend this foundation to open-ended, applied scenarios adapted from graduate examination questions. These problems require multi-step qualitative inference, synthesis of multiple concepts, and explanation under realistic constraints, approximating reasoning demands encountered in active research.

Consistent with this emphasis, MATRIX focuses on explanation-centered reasoning rather than symbolic or numerical computation. This design allows evaluation of reasoning quality independently of numerical accuracy and facilitates analysis of whether multimodal grounding influences text-only scientific reasoning.

\section{Hypothesis Generation Tasks}
Hypothesis generation tasks are constructed from research papers using restricted early-paper context (e.g., abstract and introduction), preventing leakage of reported results while encouraging reasoning from motivations, assumptions, and open challenges. Papers are filtered using domain-specific keywords.

Because multiple hypotheses may be scientifically valid, MATRIX uses a structured response format that separates problem framing, mechanistic reasoning, and the proposed hypothesis. Below is a sample hypothesis that is based on an open-access materials science paper~\cite{iioka2024evaluation}.

\begin{quote}
\textbf{QUESTION:} \\
What combination of CNF concentration and hardening alloying element addition to the nickel electroplating bath would most effectively enhance both surface hardness and subsurface mechanical stability to maximize wear resistance under high contact stress conditions? \\

\textbf{PROBLEM:} \\
The hardness enhancement mechanism achieved through Ni crystallite refinement by fine CNF co-deposition may be insufficient to prevent subsurface deformation and long-term wear under severe contact conditions, despite the surface lubrication effect of CNFs. \\

\textbf{REASONING:} \\
CNF co-deposition refines Ni crystallite size, increasing hardness through grain boundary pinning or increased nucleation during electrodeposition. However, surface hardness alone may not adequately suppress subsurface plastic deformation that governs wear progression. Incorporating hardening alloying elements such as phosphorus or tungsten, which promote solid-solution or precipitation strengthening in Ni matrices, could synergistically enhance subsurface mechanical stability while retaining the lubricating and grain-refining effects of CNFs. \\

\textbf{HYPOTHESIS:} \\
Co-deposition of an optimized CNF concentration with hardening alloying elements such as phosphorus or tungsten in the nickel electroplating bath will produce a composite coating exhibiting enhanced surface hardness and improved subsurface mechanical stability through combined grain refinement, solid-solution strengthening, and CNF-induced lubrication, thereby maximizing wear resistance under high contact stress conditions.
\end{quote}

Evaluation allows multiple acceptable answers provided they are mechanistically grounded and consistent with the prompt context. Scoring rubrics are described in~\autoref{sec:appendix_scoring}.

% \section{Multimodal Data Selection}
% \label{appendix:multimodal_data_selection}

% Reflecting real-world analysis, each modality included in MATRIX captures a distinct type of experimental evidence commonly used in materials science reasoning. SEM secondary-electron (SEM-SE) images emphasize surface morphology and topography, while backscattered-electron (SEM-BSE) images highlight compositional contrast through atomic-number sensitivity. Energy-dispersive spectroscopy (EDS) maps provide a spatially resolved elemental composition, typically interpreted jointly with microstructural features. Thermogravimetric analysis (TGA) curves encode mass-loss behavior as a function of temperature, supporting reasoning about thermal stability and decomposition pathways. X-ray diffraction (XRD) patterns capture crystallographic structure and phase composition via peak positions, intensities, and shapes.

% Interpreting these signals requires domain knowledge beyond visual description, such as linking contrast to composition, peak shifts to phase transitions, or mass-loss steps to chemical processes. The examples below illustrate the types of experimental artifacts and enriched captions used in MATRIX.

\section{Experimental Interpretation Tasks}

Experimental interpretation tasks require models to reason over real characterization data by integrating visual evidence with materials science knowledge. Here, we explain the reasoning for the kinds of figures included in the experimental interpretation tasks.

For SEM, MATRIX distinguishes between secondary-electron (SEM-SE) and backscattered-electron (SEM-BSE) imaging, requiring models to reason appropriately about surface morphology versus compositional contrast. XRD tasks require reasoning over diffraction patterns to infer phase composition and structural changes. EDS tasks involve spatially resolved elemental maps interpreted jointly with microstructural features. TGA tasks require models to reason about mass-loss curves to infer decomposition pathways and thermal stability.

\begin{figure}[t]
    \centering
    \includegraphics[width=\columnwidth]{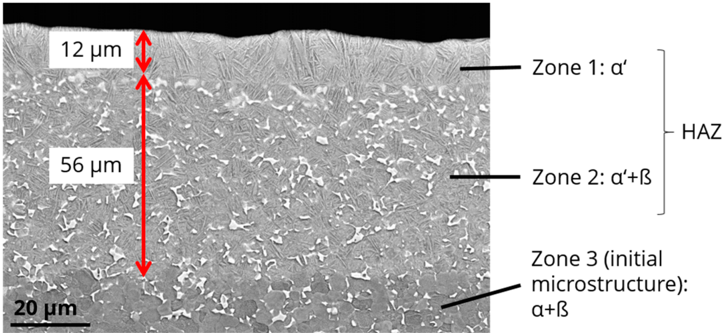}
     \begin{tcolorbox}[
          width=\columnwidth,
          colback=gray!5,
          colframe=black!60,
          title= Extracted Caption,
          fonttitle=\bfseries\footnotesize,
          sharp corners,
          boxrule=0.5pt,
          left=4pt,right=4pt,top=4pt,bottom=4pt
        ]
        \footnotesize
        Figure 5. SEM image (BSE cross-section) of the HAZ at the lower cutting edge of a Ti-6Al\_LC sample showing three distinctive microstructural zones after laser-cutting consisting of martensitic $\alpha$-phase and retained $\beta$-phase. 
        %Metals 2019 , 9 , x FOR PEER REVIEW 8  of  14
    \end{tcolorbox}
    
    \begin{tcolorbox}[
          width=\columnwidth,
          colback=gray!5,
          colframe=black!60,
          title= Prompt Context,
          fonttitle=\bfseries\footnotesize,
          sharp corners,
          boxrule=0.5pt,
          left=4pt,right=4pt,top=4pt,bottom=4pt
        ]
        \footnotesize
    
        The paper this image came from examines the microstructural evolution of alloy samples after laser cutting. It specifically characterizes the layers within the heat-affected zone (HAZ) near the cut edge, where the initial alpha and beta microstructure undergoes phase transformation into martensitic alpha'.
    
    \end{tcolorbox}
    \begin{tcolorbox}[
          width=\columnwidth,
          colback=gray!5,
          colframe=black!60,
          title= Description Caption,
          fonttitle=\bfseries\footnotesize,
          sharp corners,
          boxrule=0.5pt,
          left=4pt,right=4pt,top=4pt,bottom=4pt
        ]
        \footnotesize
    
        % {"paper_id": "W2966355370", "image_id": "page_7_img_8", "pdf_path": "/data/curtis/papers/vision_captions/W2966355370.pdf",
    
        SEM image of a sample cross-section showing three distinct microstructural regions. Zone 1 at the top surface is identified as the alpha' phase and has a measured thickness of 12 micrometers. Zone 2, located below Zone 1, consists of alpha' and beta phases with a thickness of 56 micrometers. These two top zones are collectively labeled as the Heat Affected Zone (HAZ). Zone 3, representing the initial microstructure at the bottom, is labeled alpha + beta. Red double-headed arrows indicate the depth of each measured zone. A 20 micrometer scale bar is provided in the bottom left corner.
    \end{tcolorbox}

    \caption{\footnotesize Example SEM-BSE image–caption pair and enriched description generated from the source corpus~\cite{reck2019fatigue}.}
    \label{fig:sem_bse_extracted}
\end{figure}

\begin{figure}[t]
    \centering
    \includegraphics[width=\columnwidth]{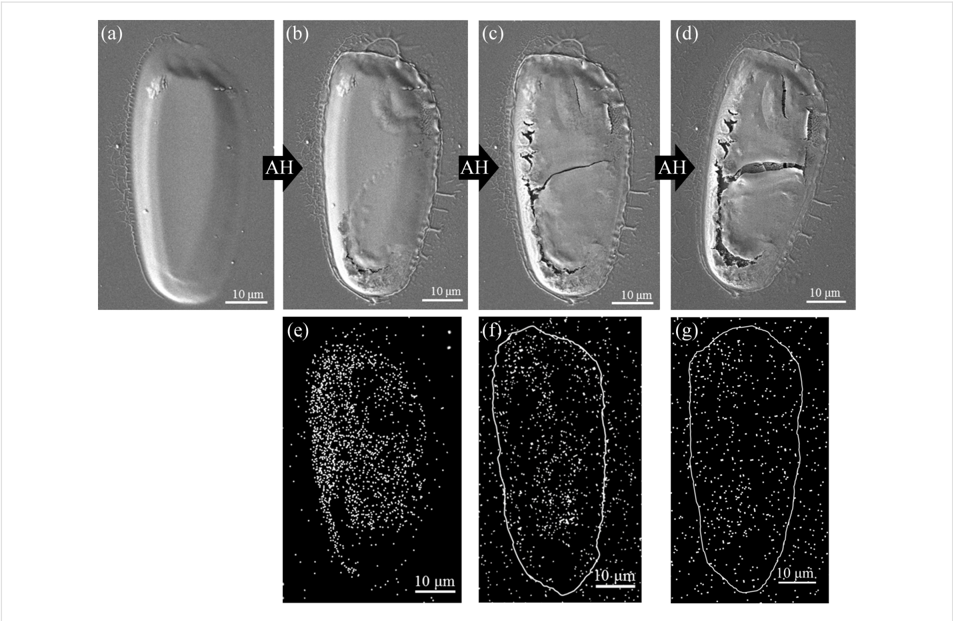}

     \begin{tcolorbox}[
      width=\columnwidth,
      colback=gray!5,
      colframe=black!60,
      title=Context,
      fonttitle=\bfseries\footnotesize,
      sharp corners,
      boxrule=0.5pt,
      left=4pt,right=4pt,top=4pt,bottom=4pt
    ]
    \footnotesize
    
    The paper this image came from describes the effect of atomic hydrogen (AH) treatment on deposits created from platinum-containing precursors. It specifically focuses on how this treatment affects the structure and chlorine content of the deposits.
    
    \end{tcolorbox}

    \begin{tcolorbox}[
      width=\columnwidth,
      colback=gray!5,
      colframe=black!60,
      title= Description Caption,
      fonttitle=\bfseries\footnotesize,
      sharp corners,
      boxrule=0.5pt,
      left=4pt,right=4pt,top=4pt,bottom=4pt
    ]
    \footnotesize
    Microscopy images and elemental maps illustrating structural and compositional changes in a deposit over the course of a treatment labeled 'AH'. (a)-(d) Sequential SEM micrographs of the deposit: (a) in its initial state and (b)-(d) following successive stages of treatment. As the treatment proceeds, the deposit develops surface textures and distinct cracks. (e)-(g) Corresponding elemental distribution maps for the stages shown in (b), (c), and (d), respectively, depicting a progressive decrease in the density of the mapped chemical species. White outlines in (f) and (g) are used to indicate the location and boundary of the deposit. A 10 µm scale bar is provided in each panel.
    
    \end{tcolorbox}

    \caption{\footnotesize Example SEM-SE image–caption pair and enriched description generated from the source corpus~\cite{spencer2017comparing}.}
    \label{fig:sem_se_extracted}

\end{figure}
%{"paper_id": "W2770935594", "image_id": "page_8_img_5", "pdf_path": "/data/curtis/papers/vision_captions/W2770935594.pdf", 

\begin{figure}[t]
    \centering
    \includegraphics[width=\columnwidth]{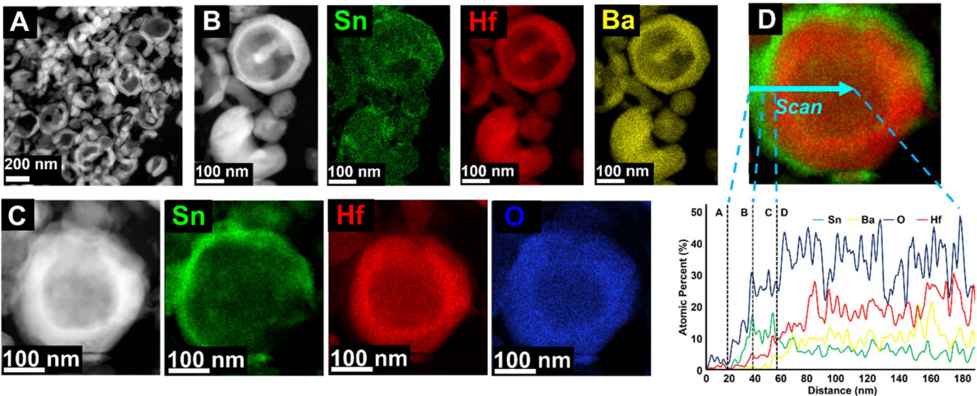}

     \begin{tcolorbox}[
      width=\columnwidth,
      colback=gray!5,
      colframe=black!60,
      title=Context,
      fonttitle=\bfseries\footnotesize,
      sharp corners,
      boxrule=0.5pt,
      left=4pt,right=4pt,top=4pt,bottom=4pt
    ]
    \footnotesize
    The paper describes the synthesis of hollow-core nanoparticles through an ion-exchange process involving Sn, Hf, Ba, and O, which results in a shell-over-nanoshell morphology.", "assumptions": "The imaging technique is identified as scanning transmission electron microscopy (STEM) based on the image characteristics and context. The elemental maps and line scan plots are interpreted based on the labels Sn (tin), Hf (hafnium), Ba (barium), and O (oxygen) provided in the figure.
    
    \end{tcolorbox}
    \begin{tcolorbox}[
      width=\columnwidth,
      colback=gray!5,
      colframe=black!60,
      title= Description Caption,
      fonttitle=\bfseries\footnotesize,
      sharp corners,
      boxrule=0.5pt,
      left=4pt,right=4pt,top=4pt,bottom=4pt
    ]
    \footnotesize

    STEM images and EDS elemental mapping of hollow nanoparticles. (A) A micrograph of several hollow nanoparticles with a 200 nm scale bar. (B) A close-up STEM image of a few nanoparticles accompanied by EDS maps showing the distribution of Sn (green), Hf (red), and Ba (yellow); scale bars are 100 nm. (C) High-magnification STEM image of a single hollow nanoparticle with its corresponding EDS maps for Sn (green), Hf (red), and O (blue); scale bars are 100 nm. (D) A composite EDS map with a scan line (blue arrow) indicating the path of an EDS line scan across a particle. The plot below displays the atomic percentage of Sn (green), Ba (yellow), O (blue), and Hf (red) versus distance in nanometers across labeled regions A, B, C, and D.
    
    \end{tcolorbox}

    \caption{\footnotesize Example EDS image–caption pair and enriched description generated from the source corpus~\cite{gabilondo2022circumventing}.}
    \label{fig:sample_eds_pair}
\end{figure}
%{"paper_id": "W2610223010", "image_id": "page_3_img_3", "pdf_path": "/data/curtis/papers/vision_captions/W2610223010.pdf", 

\begin{figure}[t]
    \centering
    \includegraphics[width=\columnwidth]{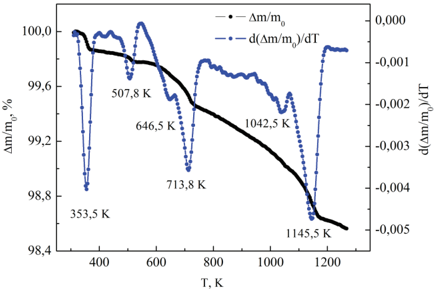}

    \begin{tcolorbox}[
      width=\columnwidth,
      colback=gray!5,
      colframe=black!60,
      title=Context,
      fonttitle=\bfseries\footnotesize,
      sharp corners,
      boxrule=0.5pt,
      left=4pt,right=4pt,top=4pt,bottom=4pt
    ]
    \footnotesize
    The paper this image came from investigates the thermal synthesis of double perovskite materials starting from a stoichiometric mixture of SrCO3, Cr2O3, and MoO3. The synthesis process involved heating the powder mixture from 300 to 1300 K in a continuous 5\% H2/Ar gas flow at a rate of 1.4 K/min.

    \end{tcolorbox}

    \begin{tcolorbox}[
      width=\columnwidth,
      colback=gray!5,
      colframe=black!60,
      title= Description Caption,
      fonttitle=\bfseries\footnotesize,
      sharp corners,
      boxrule=0.5pt,
      left=4pt,right=4pt,top=4pt,bottom=4pt
    ]
    \footnotesize
    A dual-axis plot showing the normalized mass change and its temperature derivative for a powder sample as a function of temperature (T) in Kelvin, ranging from 300 to 1300 K. The solid black line with circular markers represents the normalized mass change (delta m/m0, \%), with values indicated on the left vertical axis showing a stepwise decrease in mass. The blue dashed line with circular markers represents the derivative of the normalized mass change $\frac{d}{dT}\left(\frac{\Delta m}{m_0}\right)$ on the right vertical axis. Significant mass loss events are highlighted by minima in the derivative curve, which are labeled with specific temperatures: 353.5 K, 507.8 K, 646.5 K, 713.8 K, 1042.5 K, and 1145.5 K. These measurements were obtained while heating the sample at a rate of 1.4 K/min in a 5\% H2/Ar environment.

    \end{tcolorbox}

    \caption{\footnotesize Example TGA image–caption pair and enriched description generated from the source corpus~\cite{kalanda2019sequence}.}
    \label{fig:sample_tga_pair}
\end{figure}
%{"paper_id": "W3010748023", "image_id": "page_3_img_1", "pdf_path": "/data/curtis/papers/vision_captions/W3010748023.pdf",

\begin{figure}[t]
    \centering
    \includegraphics[width=\columnwidth]{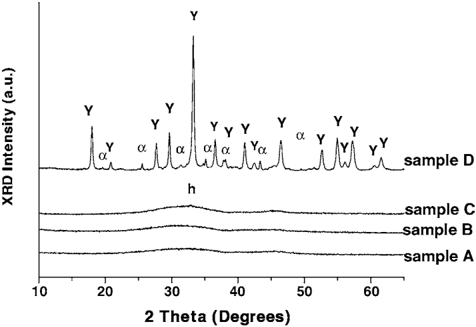}

    \begin{tcolorbox}[
      width=\columnwidth,
      colback=gray!5,
      colframe=black!60,
      title=Context,
      fonttitle=\bfseries\footnotesize,
      sharp corners,
      boxrule=0.5pt,
      left=4pt,right=4pt,top=4pt,bottom=4pt
    ]
    \footnotesize
    The samples (A, B, C, and D) were prepared with different pre-treatment temperatures: sample A at 600 C for 30 min, sample B at 800 C for 30 min, sample C at 900 C for 30 min, and sample D at 1215 C.

    \end{tcolorbox}

    \begin{tcolorbox}[
      width=\columnwidth,
      colback=gray!5,
      colframe=black!60,
      title= Description Caption,
      fonttitle=\bfseries\footnotesize,
      sharp corners,
      boxrule=0.5pt,
      left=4pt,right=4pt,top=4pt,bottom=4pt
    ]
    \footnotesize

    XRD intensity patterns as a function of the 2-theta angle (10–65 degrees) for four samples: sample A, sample B, sample C, and sample D. Samples A, B, and C show broad, low-intensity features with a hump labeled 'h' around 33 degrees. Sample D displays multiple sharp diffraction peaks, which are labeled with 'Y' and 'alpha' to indicate specific crystalline phases.

    \end{tcolorbox}

    \caption{\footnotesize Example XRD image–caption pair and enriched description generated from the source corpus~\cite{palmero2006comparison}.}
    \label{fig:sample_xrd_pair}
\end{figure}
%{"paper_id": "W2032776929", "image_id": "page_4_img_4", "pdf_path": "/data/curtis/papers/vision_captions/W2032776929.pdf",

\begin{figure}[t]
    \centering
    \includegraphics[width=\columnwidth]{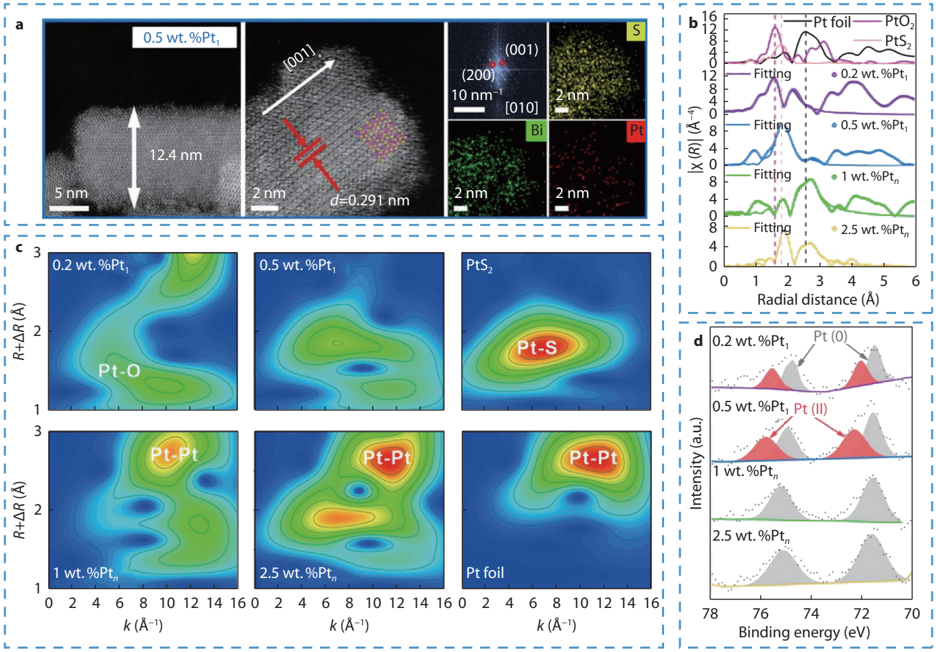}
    
    \begin{tcolorbox}[
      width=\columnwidth,
      colback=gray!5,
      colframe=black!60,
      title=Context,
      fonttitle=\bfseries\footnotesize,
      sharp corners,
      boxrule=0.5pt,
      left=4pt,right=4pt,top=4pt,bottom=4pt
    ]
    \footnotesize
    The paper this image came from discusses the synthesis and characterization of single-atom platinum (Pt1) catalysts supported on bismuth sulfide (Bi2S3) nanowires for thermoelectric applications. It compares samples with individual Pt atoms (Pt1) to those with Pt nanoclusters (Ptn)

    \end{tcolorbox}

    \begin{tcolorbox}[
      width=\columnwidth,
      colback=gray!5,
      colframe=black!60,
      title= Description Caption,
      fonttitle=\bfseries\footnotesize,
      sharp corners,
      boxrule=0.5pt,
      left=4pt,right=4pt,top=4pt,bottom=4pt
    ]
    \footnotesize

    (a) Scanning transmission electron microscopy (STEM) images of a Bi2S3 sample with 0.5 wt.\% Pt1. The panel includes a low-magnification image showing a thickness of 12.4 nm, a high-resolution lattice image with a d-spacing of 0.291 nm along the [001] direction and a superimposed crystal structure model, a corresponding diffraction pattern, and energy-dispersive X-ray spectroscopy (EDS) elemental maps for bismuth (Bi, green), sulfur (S, yellow), and platinum (Pt, red). (b) Fourier transform extended X-ray absorption fine structure (FT-EXAFS) spectra for Pt foil, PtO2, PtS2, and various Pt-decorated Bi2S3 samples, with vertical dashed lines marking key coordination shell positions at approximately 1.6 Å and 2.6 Å. (c) Wavelet transform (WT) plots of the EXAFS signals identifying coordination environments such as Pt-O, Pt-S, and Pt-Pt. (d) X-ray photoelectron spectroscopy (XPS) spectra of the Pt 4f region showing a transition from Pt(II) oxidation states (red) to a metallic Pt(0) state (gray) as the Pt concentration increases from 0.2 to 2.5 wt.\%.

    \end{tcolorbox}

    \caption{\footnotesize Example other image–caption pair and enriched description generated from the source corpus~\cite{huang2023single}.}
    \label{fig:sample_other_pair}
\end{figure}

%{"paper_id": "W4352990983", "image_id": "page_2_img_2", "pdf_path": "/data/curtis/papers/vision_captions/W4352990983.pdf",

\section{Scoring}
\label{sec:appendix_scoring}

MATRIX tasks are judged using a five-level scale because scientific explanations often vary along multiple dimensions, such as correctness of conclusions, completeness of mechanistic reasoning, and alignment with experimental evidence. In preliminary experiments, coarser scoring schemes collapsed qualitatively distinct responses into the same category, reducing sensitivity to partial improvements.

The judge is provided with the task prompt, a reference answer or rubric, and the model-generated response. For hypothesis-generation tasks, evaluation emphasizes clarity of problem framing, quality of mechanistic reasoning, and plausibility of the proposed hypothesis rather than adherence to a single expected answer. For experimental interpretation tasks, scoring prioritizes identification of salient features and their correct linkage to physical mechanisms over surface-level description.

To assess judge stability, a stratified subset of responses is independently re-scored using a different judge model, yielding consistent relative rankings. The judge model is not used to generate any training supervision for evaluated models, and none of the enhanced image captions used during post-training overlap with judge inputs. 

\subsection{Hypothesis Block Schema}
\label{prompt_hypothesis_structure}

    \begin{tcolorbox}[
      width=\columnwidth,
      colback=gray!5,
      colframe=black!60,
      title= Scoring hypothesis Schema,
      fonttitle=\bfseries\footnotesize,
      sharp corners,
      boxrule=0.5pt,
      left=4pt,right=4pt,top=4pt,bottom=4pt
    ]
    \footnotesize
    
    - \textbf{1.0: Excellent} — Clear problem framing, scientifically plausible and well-grounded reasoning, and a specific, testable hypothesis directly tied to the reasoning.\\
    - \textbf{0.75: Good} — Generally clear and plausible; minor gaps, vagueness, or missing details in either reasoning or hypothesis.\\
    - \textbf{0.5: Partial} — Some correct ideas or partial framing, but weak or incomplete scientific grounding and/or hypothesis not clearly testable.\\
    - \textbf{0.25: Poor} — Minimal structure; vague or generic problem, shallow or loosely related reasoning, and unclear hypothesis.\\
    - \textbf{0.0: Incorrect} — Scientifically implausible, factually wrong, or irrelevant to the question.\\

    \end{tcolorbox}

\section{Post-Training Data and Experimental Setup}
\label{sec:appendix_post_training}

This section provides detailed statistics and implementation details for the post-training experiments described in the main paper.

\subsection{Post-Training Data Composition}

Post-training supervision spans four task families: foundational theory reasoning, research-level reasoning, hypothesis generation, and experimental interpretation. Text-based supervision is drawn from postgraduate coursework and research-derived prompts, while multimodal supervision consists of aligned experimental image–description pairs.

Foundational theory comprises 8,100 training questions and 900 validation questions. These questions focus on core materials science concepts such as thermodynamics, crystallography, phase equilibria, and structure–property relationships.

Research-level reasoning consists of 1,138 training questions and 126 validation questions. These tasks are adapted from graduate examination and research-style prompts and require multi-step qualitative inference, synthesis of multiple concepts, and explicit discussion of assumptions or trade-offs.

Hypothesis-generation supervision includes 1,442 training prompts and 161 validation prompts. These tasks are constructed from early sections of research papers (e.g., abstracts and introductions) and require proposing mechanistically grounded hypotheses consistent with the provided context.

Multimodal supervision comprises 3,056 aligned experimental image–text pairs with 760 validation pairs. Image-based tasks span SEM secondary-electron (SEM-SE), SEM backscattered-electron (SEM-BSE), XRD, EDS, TGA, and an auxiliary “other” category. For evaluation, 50 held-out test examples are used per experimental modality, yielding 300 total evaluation examples across text-based and image-based tasks.

\subsection{Train–Validation–Test Separation and Contamination Control}

To prevent contamination between post-training and evaluation, all datasets are partitioned at the document level. For image-based and hypothesis-generation supervision, post-training examples are drawn exclusively from research papers disjoint from those used to construct evaluation tasks.

For text-based tasks, foundational theory and research reasoning questions are expanded and filtered to ensure no overlap in prompt content, structure, or source materials with held-out evaluation questions. No figures, captions, or textual passages from evaluation documents are included in post-training supervision.

All post-training configurations use the same held-out evaluation set, and no post-training data are reused for scoring or rubric construction. This separation follows standard practices in domain-adaptive pretraining and controlled fine-tuning experiments.

\subsection{Scale and Experimental Intent}

The overall scale of post-training is intentionally modest relative to large instruction-tuning corpora. MATRIX tasks require long-form, mechanistically grounded responses evaluated using rubric-guided LLM judging, which introduces nontrivial cost and variance at scale. Consequently, we prioritize diagnostic sensitivity and controlled comparison over leaderboard-scale evaluation.

The data volumes used here are sufficient to surface consistent and interpretable differences across post-training configurations, enabling analysis of how different supervision signals—textual reasoning versus aligned visual grounding—affect scientific reasoning behavior.

\section{Permutation Procedure}
\label{sec:appendix_permutation}

Images (and captions) are permuted in an anti-stratified manner, where images (e.g. a TGA image) are replaced by an image from a different kind of question (e.g. EDS). This ensures that the permuted image/caption must be incorrect. All variants are trained using the same data sources, architecture, and optimization settings, differing only in how image–text alignment is structured.

\section{Additional Results}
\label{sec:appendix_add_res}

\begin{table*}[t]
\centering
\small
\setlength{\tabcolsep}{8pt}
\renewcommand{\arraystretch}{1.15}
\begin{tabular}{lccc}
\toprule
\textbf{Model} & \textbf{Theory} & \textbf{Research Reasoning} & \textbf{Hypothesis} \\
\midrule
% \multicolumn{4}{l}{\textit{Domain-Specialized Text-Only Baselines}} \\

\multicolumn{4}{l}{\textit{Text-Only Post-Training}} \\
Finetuned on Foundational Theory & 0.435 & 0.575 & 0.555 \\
Finetuned on Research Reasoning & 0.450 & 0.550 & 0.530 \\
Finetuned on Hypothesis Questions & 0.435 & 0.620 & 0.520 \\
Finetuned on Theory, Reasoning, and Questions & 0.470 & 0.580 & 0.565 \\
\bottomrule
\end{tabular}
\caption{Performance comparison of domain-specialized baselines and task-specific post-training approaches on text-based materials science reasoning tasks. LLaMaT models are continually pre-trained on large materials science corpora, while our models use targeted supervised fine-tuning on specific reasoning tasks.}
\label{tab:text_only_comparison}
\end{table*}

\autoref{tab:text_only_comparison} compares task-specific text-only post-training strategies across three materials science reasoning dimensions. Fine-tuning on foundational theory yields strong and consistent improvements across all tasks, indicating that conceptual grounding in core principles transfers effectively to both research-level reasoning and hypothesis generation. This suggests that emphasizing theoretical fundamentals provides a broad inductive bias that supports downstream scientific explanation.

Fine-tuning exclusively on hypothesis-generation tasks produces the strongest performance on research reasoning (0.620), surpassing even the combined multi-task model. This result highlights that hypothesis-generation prompts implicitly demand deep research-level reasoning, including problem framing, mechanistic justification, and consideration of alternative explanations. As a result, such tasks appear to be a particularly efficient form of supervision for developing advanced scientific reasoning capabilities.

Combining all three task types yields the most balanced overall performance, achieving strong scores across theory, research reasoning, and hypothesis generation. This indicates that while individual task types emphasize different aspects of scientific reasoning, multi-task post-training integrates complementary skills without substantial interference.

Overall, these results demonstrate that carefully designed, task-specific supervised fine-tuning on long-form reasoning problems can substantially shape scientific reasoning behavior even in the absence of visual input. The findings underscore the importance of reasoning curriculum design—rather than raw data scale or domain exposure alone—for developing robust text-only scientific reasoning in foundation models.

%%%%%%%%%%%%%%%%%%%%%%%%%%%%%%%%%%%%%%%%%%%%%%%%%%%%%%%%%%%%%%%%%%%%%%%%%%%%%%%
%%%%%%%%%%%%%%%%%%%%%%%%%%%%%%%%%%%%%%%%%%%%%%%%%%%%%%%%%%%%%%%%%%%%%%%%%%%%%%%

\end{document}